\documentclass{article}

\usepackage{microtype}
\usepackage{graphicx}
\usepackage{subcaption}
\usepackage{booktabs}
\usepackage{hyperref}

\usepackage[accepted]{icml2026}

\usepackage{amsmath}
\usepackage{amssymb}
\usepackage{mathtools}
\usepackage{amsthm}
\usepackage[capitalize,noabbrev]{cleveref}

\theoremstyle{plain}

\theoremstyle{definition}

\theoremstyle{remark}

\icmltitlerunning{Vision-Language Asymmetry in Bistable Image Captioning}

\begin{document}

\twocolumn[
  \icmltitle{Vision-Language Asymmetry in Bistable Image Captioning}

  \icmlsetsymbol{equal}{*}
  \begin{icmlauthorlist}
    \icmlauthor{Arohan Agate}{uw}
  \end{icmlauthorlist}
  \icmlaffiliation{uw}{University of Washington}
  \icmlcorrespondingauthor{Arohan Agate}{aagate@cs.washington.edu}

  \icmlkeywords{aspect-seeing, Wittgenstein, vision-language models, sparse autoencoders,
    philosophy of perception, mechanistic interpretability}

  \vskip 0.3in
]

\printAffiliationsAndNotice{}

\begin{abstract}
Wittgenstein's duck--rabbit poses a question for vision-language models: when a model captions an ambiguous image, where in the model is the commitment to one aspect made? We address this with a 3{,}320-generation behavioral baseline over 83 bistable stimuli that surfaces three regimes (\emph{default-dominant}, \emph{force-dominant}, \emph{force-balanced}) under neutral vs forced-choice prompting, then probe the underlying representations using a TopK sparse autoencoder we train on the CLIP layer that LLaVA-1.6-7B actually consumes (validation EV~0.93). Across 69 bistable stimuli with both per-aspect feature pools available, 72\% (50/69) show simultaneous activation of both pools at the vision tower, including 12/12 default-dominant duck/rabbit and 7/8 force-balanced young/old. Causal steering at CLIP layer 22 flips captions on default-dominant stimuli (33\% rabbit-flip rate under a fluency guard) but cannot flip captions on force-balanced young/old at any tested coefficient, despite their vision-side superposition. The dominance bottleneck lives downstream of the vision tower; the gap between vision-side representation and language-side commitment is an empirical handle on the seeing/seeing-as distinction. We also flag a methodological note: rank-based statistics on TopK SAE outputs require tie-corrected ranking to avoid silent row-order bias.
\end{abstract}

\section{Introduction}
\label{sec:intro}

A Necker cube shown to LLaVA-1.6-Vicuna-7B \citep{liu2024improved} under the prompt ``What is in this image?'' yields, across 40 sampled generations, captions like ``a wireframe drawing of a cube'' that commit to neither face-up nor face-down, 40/40 aspect-agnostic. ``Is this a cube viewed from above or from below?'' on the same image yields a near-even split between two committed answers. Same input, different prompt; same model, different report. The difference is exactly the difference Wittgenstein flagged in the late \emph{Philosophical Investigations} \citep[Philosophy of Psychology --- A Fragment, §§111--136]{wittgenstein2009pi}: between \emph{seeing} an image and \emph{seeing-as}. This paper takes that prompt contrast as an empirical handle on the distinction and asks where in a VLM the gap between the two reports is implemented.


We surface three behavioral regimes, \emph{default-dominant}, \emph{force-dominant}, \emph{force-balanced}, across 83 bistable stimuli and 3{,}320 generations. On 50/69 (72\%) bistable stimuli, per-aspect SAE feature pools at LLaVA's consumption layer are simultaneously active. Causal steering at CLIP layer~22 flips captions on default-dominant duck/rabbit (33\% rabbit-flip at $\alpha=16$ under a fluency guard) but \emph{cannot} flip captions on force-balanced young/old at any $\alpha$, despite their vision-side superposition. The seeing-as commitment in this VLM lives downstream of the vision tower.


\section{Background and related work}
\label{sec:background}

\paragraph{Bistable images in vision-language models.}
\citet{panagopoulou2024aspect} establish a 29-image bistable benchmark and show that twelve VLMs exhibit strong language-prior dominance on duck/rabbit, vase/profile, young/old, and similar paired figures. Their analysis is purely behavioral: captions are scored against canonical aspects, dominance is a per-stimulus statistic, and no representational claim is made. The recent \textsc{AmbiBench} submission \citep{ambibench2026} extends the behavioral side substantially (2{,}238 ambiguous images, broader aspect taxonomy) and crosses into mechanism with attention-head-level intervention, raising InternVL3-2B accuracy from 29\% to 42\% by amplifying perceptual-switch heads. Our work differs in granularity (SAE features rather than attention heads, with per-feature max-activating-image evidence in Appendix~\ref{app:phase2}) and in framing: \textsc{AmbiBench} measures task-accuracy improvement under intervention; we measure where representational commitment is localized.

\paragraph{Sparse autoencoders in CLIP/LLaVA pipelines.}
\citet{pach2025sparse} establish the CLIP-side SAE → LLaVA pipeline that we use, demonstrating that interventions on monosemantic SAE features in CLIP residual streams propagate cleanly to LLaVA's captions; \citet{joseph2025steering} quantify steerability and report that roughly 10--15\% of CLIP SAE features are reliably steerable. Our contribution is not the pipeline but its application to \emph{representational competition}: bistable stimuli are the minimal experimental condition under which feature competition (rather than feature detection) is the phenomenon of interest, and the comparison of vision-side superposition to language-side commitment is meaningful only on stimuli that admit two simultaneous interpretations.

\paragraph{Aspect-seeing in philosophy of perception.}
\citet{wittgenstein2009pi} (\emph{Philosophical Investigations}, Philosophy of Psychology --- A Fragment, §§111--136) is the canonical source for the seeing/seeing-as distinction; \citet{hanson1958patterns} reframes it at the level of scientific observation (what counts as \emph{seeing} a phenomenon is shaped by the conceptual frame the observer brings), and \citet{kuhn1962structure} generalizes this to paradigm-level commitments. \citet{nanay2016aesthetics} replaces the gestalt framing with an attention-based account: aspects are the contents the perceiver currently attends to within an otherwise stable perceptual field. Our experimental contrast engages most directly with the Wittgenstein--Hanson side: the prompt contrast is a controlled manipulation of the conceptual frame the model is asked to apply, and the coexistence of vision-side feature pools with split caption-level commitment is the mechanistic analogue. To our knowledge, no prior ML paper operationalizes this literature with mechanistic evidence; the closest adjacent work argues for philosophy–ML integration without empirical aspect-seeing experiments \citep{milliere2024interventionist,williams2025philosophy}.

\section{Methods}
\label{sec:methods}

\paragraph{Stimuli.}
Our bistable set comprises 83 images: 29 from \citet{panagopoulou2024aspect}, 53 from \textsc{AmbiBench} \citep{ambibench2026}, and one rendered Necker cube. We organize these into six groups with paired pure-aspect SDXL controls (hand-verified, 21--27 per aspect for four groups; \texttt{schroeder\_stairs} and \texttt{necker\_cube} pure-B counts are 8 and 15, see Appendix~\ref{app:dataset}): \texttt{duck\_rabbit}, \texttt{face\_vase}, \texttt{hidden\_face}, \texttt{young\_old\_woman}, \texttt{schroeder\_stairs}, \texttt{necker\_cube}.

\paragraph{Models.}
The target VLM is LLaVA-1.6-Vicuna-7B \citep{liu2024improved}, which sets \texttt{vision\_feature\_layer}\,$=-2$ on a CLIP ViT-L/14-336 \citep{radford2021learning} backbone with Vicuna 7B \citep{vicuna2023}. Representational analysis happens at CLIP layer~22 (the layer LLaVA consumes) on patch tokens, CLS dropped. Captions are classified by Qwen3-8B \citep{qwen3_2025} with \texttt{enable\_thinking=False}; judge--manual agreement is $\geq$\,95\% on 30 hand-checked captions per phase.

\paragraph{Sparse autoencoder.}
We train a TopK SAE ($k=32$, 65{,}536 features) on a 200K-image cache of CLIP layer-22 patch activations from CC3M, validation EV\,=\,0.93. We trained our own SAE rather than reuse pretrained CLIP-Scope \citep{ewingtonpitsos2024clipscope} because CLIP-Scope was trained on LAION-CLIP activations and on our bistable stimuli reconstructs at $\sim$40\% above its own MSE baseline; LLaVA consumes OpenAI-CLIP, so we matched that distribution.

\paragraph{Tie-corrected rank AUROC.}
TopK SAE outputs are over 99\% sparse, so most feature columns contain many activations tied at exactly zero. \texttt{numpy.argsort} is a stable sort and resolves ties by input-row position, which silently biases any rank-based statistic toward whichever class occupies the earlier rows of the activation matrix. We compute all AUROCs with \texttt{scipy.stats.rankdata(method="average")}, which assigns each tied set the average of the rank positions it spans. Without this correction an earlier pass returned roughly 14k phantom B-preferring features per group, all clustered in one feature-index range (Appendix~\ref{app:rank-tie}).

\paragraph{Phase 2 --- feature identification.}
For each group and each aspect $X\in\{A,B\}$, we compute the leave-one-out tie-corrected AUROC of every SAE feature in separating mean-pooled control activations of class $X$ from those of the opposite class. We retain features with leave-one-out AUROC $\geq 0.85$ \emph{and} mean-match activation $> 0.005$ (a sparsity floor that excludes features which fire on almost no image), then compute a second AUROC against 10K random CC3M patches as a distractor specificity test. Surviving features are ranked by distractor AUROC and capped at 15 per aspect.

\paragraph{Phase 3 --- superposition vs.\ dominance.}
For each bistable stimulus we compute the mean activation of its 15 A-features and 15 B-features, which we call the A-pool and B-pool activations. The threshold for ``$X$-pool fires'' is the median activation of $X$-pool features observed on \emph{opposite-aspect} controls, i.e.\ the activation level that opposite-class images can produce by chance. A stimulus is classified \emph{superposition} when both pools exceed their thresholds, \emph{dominance\_X} when only $X$ does, and \emph{neither} otherwise.

\paragraph{Phase 4 --- causal steering.}
The steering vector for aspect $X$ is the mean of the 15 $X$-feature SAE decoder rows. We add $\alpha v_X$ to the layer-22 patch residual under a forward hook, sweep $\alpha\in\{2^{-1},\ldots,2^4\}$, and generate one caption per stimulus per $\alpha$. Success is the fraction Qwen3 labels as the steered aspect; we accept an $\alpha$ only if perplexity ratio against the unsteered caption is $\leq$\,1.2 (the fluency guard). Steering runs only on Phase-3 superposition stimuli.

\paragraph{Reproducibility.} Code, configs, and per-stimulus result tables will be released upon acceptance.

\section{Results}
\label{sec:results}

\subsection{Three behavioral regimes}
\label{sec:results-behaviour}

\begin{figure}[!t]
  \centering
  \includegraphics[width=\columnwidth]{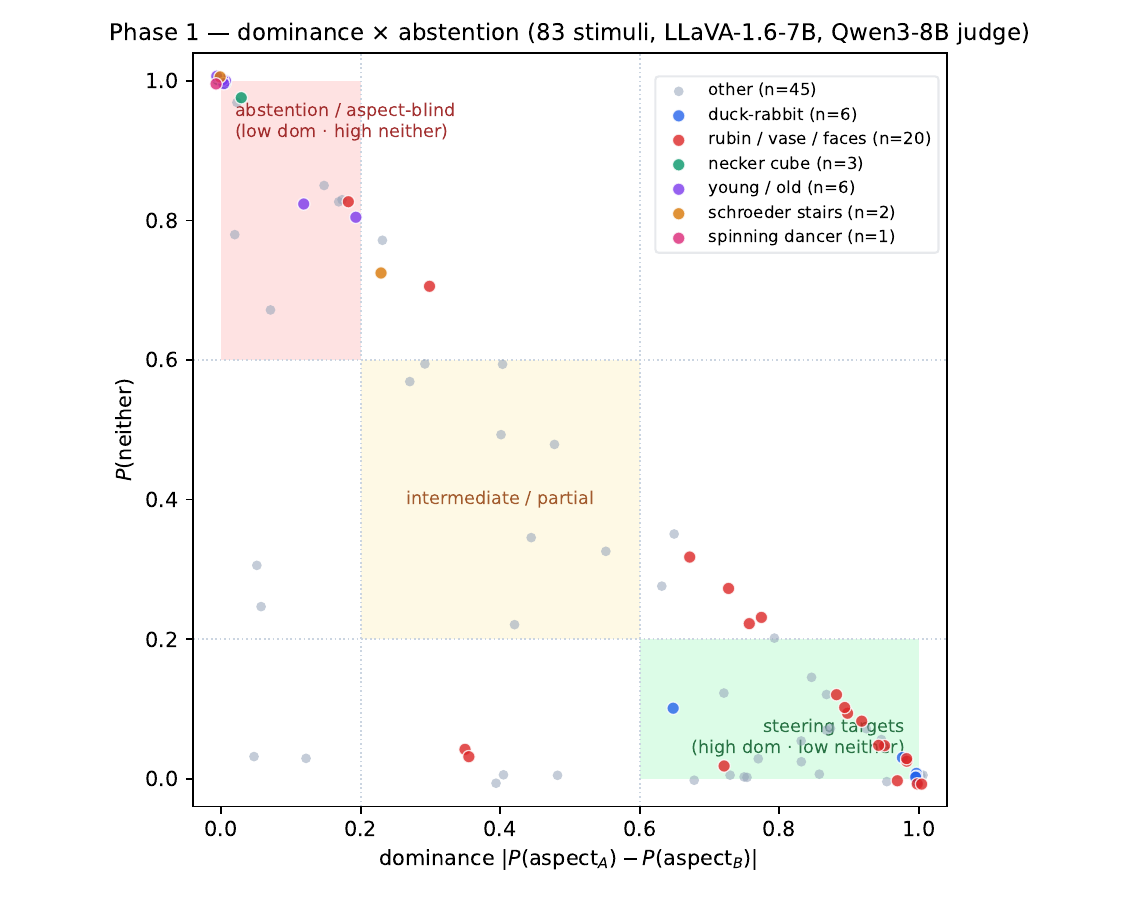}
  \caption{Three behavioral regimes over 83 bistable stimuli (LLaVA-1.6-7B, 40 generations each, Qwen3-8B judge). Bottom-right: \emph{default-dominant}; top-left: aspect-blind under neutral prompting, splits into \emph{force-dominant} (7/10, $\geq$70/30 commitment under forced choice) and \emph{force-balanced} (3/10, $\sim$50/50, the Necker cubes).}
  \label{fig:dominance}
\end{figure}

Across 3{,}320 captions over 83 stimuli (40 generations per stimulus, neutral prompts ``What is in this image?'' and ``Describe this image''), LLaVA's mean dominance score $|P(A)-P(B)|$ is 0.558 and 38 of 83 stimuli show dominance $>0.5$ with $P(\text{neither})<0.2$, the classical \emph{default-dominant} regime that replicates the language-prior bias of \citet{panagopoulou2024aspect}. Ten stimuli sit in the opposite corner of \cref{fig:dominance}: $P(\text{neither})\geq 0.95$ under neutral prompting, i.e.\ LLaVA refuses to commit to either aspect and produces aspect-agnostic descriptions (``a black-and-white line drawing of a person''). A second behavioral pass on these ten stimuli replaces the neutral prompt with a binary forced-choice prompt (``Is this an X or a Y?''). All ten stimuli (10/10) commit to an aspect under forced choice. Seven commit asymmetrically with at least a 70/30 split (\emph{force-dominant}: Schroeder stairs, spinning dancer, two of three young/old woman exemplars, two AmbiBench young/old, the grimace--beggar) and three split close to 50/50 (\emph{force-balanced}: all three Necker cube exemplars). Behavioral abstention is not representational unavailability: the model has aspect-specific information available, it simply does not surface it under neutral prompts. The three-regime taxonomy gives us three different mechanistic predictions for what should happen at the SAE-feature level, which we test in \cref{sec:results-superposition,sec:results-steering}.

\subsection{Per-aspect SAE features exist}
\label{sec:results-features}

For each of the six analysis groups, the leave-one-out tie-corrected AUROC plus mean-match floor plus distractor filter retains exactly 15 features per aspect (30 features per group, 180 features overall). Median leave-one-out AUROC is $\geq 0.997$ across all six groups; median CC3M-distractor AUROC is $\geq 0.998$. The leave-one-out split holds out every control image once, and the surviving features must separate matched against opposite-aspect controls under every fold; max-activating images cleanly partition by aspect (Appendix~\ref{app:phase2} reproduces all six groups' grids and the duck/rabbit + face/vase compact view from \cref{fig:features-app}).

\begin{figure*}[!htb]
  \centering
  \includegraphics[width=\textwidth, trim={20pt 0 0 30pt}, clip]{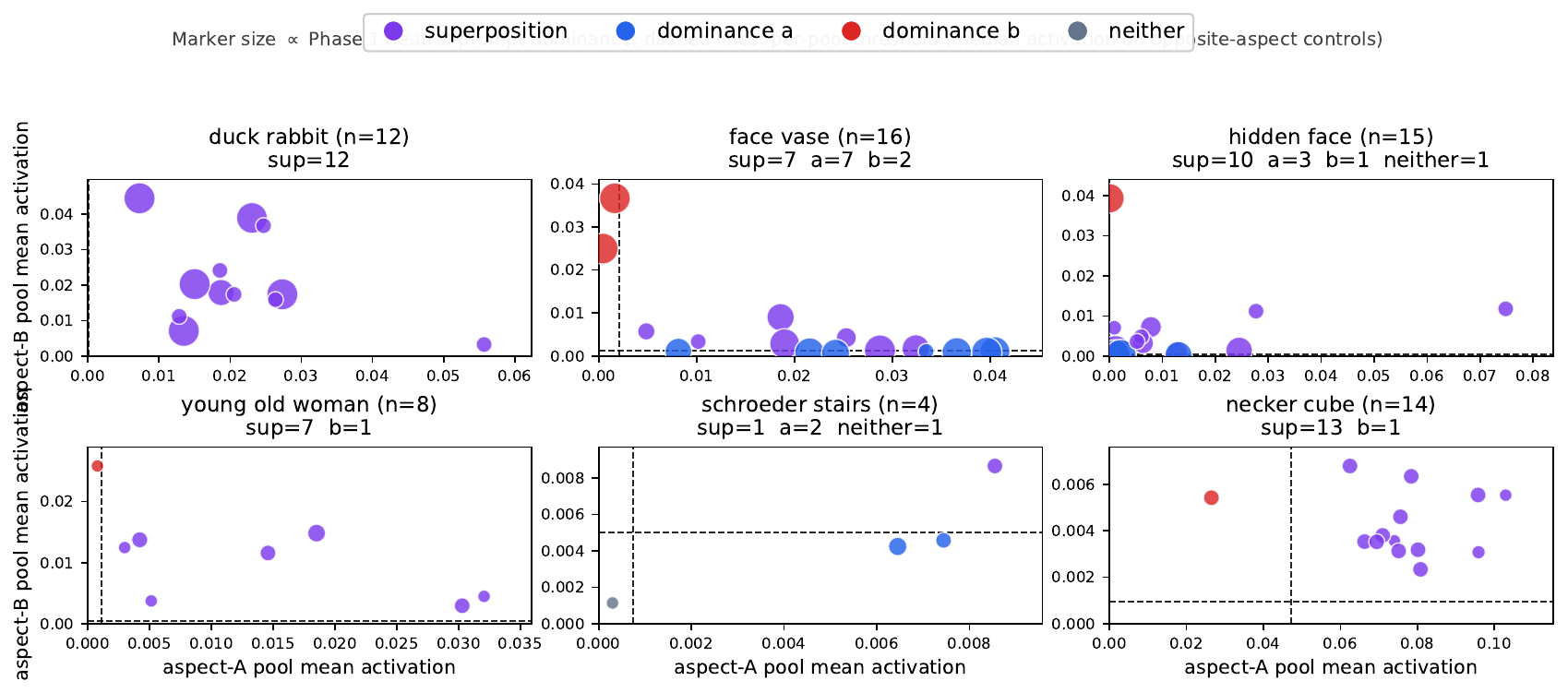}
  \caption{Phase 3: per-stimulus aspect-A (x-axis) vs aspect-B (y-axis) pool mean activation across all six groups. Marker color gives the per-stimulus classification (purple = superposition, blue = dominance A, red = dominance B, gray = neither); marker size is proportional to Phase 1 neutral-prompt dominance score. Dashed lines show per-pool thresholds (median activation on opposite-aspect controls). Purple dominates: both pools above threshold (vision-side superposition) for 50 of 69 stimuli.}
  \label{fig:superposition}
\end{figure*}
\subsection{Vision-side superposition is the modal regime}
\label{sec:results-superposition}

For each bistable stimulus we compute the mean activation of its 15 A-features and 15 B-features and compare each pool to the median activation observed for that pool on \emph{opposite}-aspect controls (\cref{sec:methods}). \cref{fig:superposition} shows the result across all six groups and all 69 bistable stimuli for which both pools are available. Fifty stimuli (50/69, 72\%) classify as \emph{superposition}: both feature pools fire above their opposite-class threshold. The per-group breakdown: \texttt{duck\_rabbit} 12/12, \texttt{necker\_cube} 13/14, \texttt{young\_old\_woman} 7/8, \texttt{hidden\_face} 10/15, \texttt{face\_vase} 7/16, and \texttt{schroeder\_stairs} 1/4. The Schroeder count is the weakest case: pure-B controls were thin after curation ($n=8$), and we report Schroeder for completeness of the figure; per-stimulus details and the curation log are in Appendix~\ref{app:dataset}. The result that matters for the rest of the paper is that vision-side superposition is the \emph{modal} regime regardless of behavioral outcome: it is unanimous on default-dominant duck--rabbit ($n=12$, all duck-leaning behaviorally), nearly unanimous on the force-balanced Necker cubes (13/14), and 7/8 on the force-dominant young--old. Whatever distinguishes these three behavioral regimes downstream, it is not happening at the SAE feature level of CLIP layer~22.

\subsection{Causal steering exposes a vision/language asymmetry}
\label{sec:results-steering}

Phase 4 takes each Phase-3 superposition stimulus, constructs the steering vector for the non-default aspect as the mean of the 15 target-aspect SAE decoder rows, injects it into CLIP layer~22, sweeps $\alpha$, and measures both caption-flip rate and the perplexity ratio against the unsteered caption (\cref{sec:methods}). We run three groups, one per behavioral regime.

For the \emph{default-dominant} group \texttt{duck\_rabbit} (baseline 91.7\% duck), steering toward duck is a no-op (already at the behavioral ceiling); steering toward rabbit reaches 33.3\% rabbit captions at $\alpha=16$ with perplexity ratio 1.06, comfortably under the 1.2$\times$ fluency guard. For the mixed group \texttt{hidden\_face} (baseline 30\% A / 50\% B / 10\% neither), steering toward A reaches 60\% at $\alpha=16$ (perplexity ratio 0.99) and steering toward B reaches 50\% at $\alpha=8$ (1.04). For the \emph{force-balanced} group \texttt{young\_old\_woman} (baseline 0\% A / 0\% B / 100\% neither), \emph{neither} direction produces a single aspect-committed caption at any $\alpha$ tested, despite Phase 3 finding 7/8 of these stimuli in vision-side superposition.\footnote{Phase 4 steers only the seven \texttt{young\_old\_woman} stimuli Phase 3 classified as superposition; the eighth (\texttt{pana\_028}) was \texttt{dominance\_b} and excluded by the protocol's superposition filter.} The steering direction visibly affects low-level visual descriptors in the captions (hair, hat, lighting): at $\alpha{=}16$ the captions describe ``a stylized illustration of a woman's face with a dramatic, exaggerated eyelash and a large flowing hair accessory that resembles a feather or a piece of fabric'' -- the steering direction is reaching the language model, but the language model continues to refuse to commit to ``young'' or ``old.'' Captions remain aspect-agnostic across the entire $\alpha$ range and degrade fluency before flipping (\cref{app:phase4}).

\section{Discussion}
\label{sec:discussion}


\paragraph{Operationalizing seeing/seeing-as without metaphysical commitment.}
Wittgenstein's distinction between \emph{seeing} an image and \emph{seeing-as} was a remark on a structural feature of perception, not a claim about phenomenology \citep{wittgenstein2009pi}. Our prompt contrast inherits the structure without the phenomenology: the neutral prompt asks the model to report what it sees, the forced-choice prompt to commit to seeing-as. The behavioral gap (100\% abstention becoming 100\% commitment on ten stimuli) is a measurable instance of the gap Wittgenstein flagged. We do not claim LLaVA \emph{has} aspect-seeing experiences; we claim the prompt contrast tracks the same distinction in a system whose internal state we can inspect.
\paragraph{Theory-ladenness as a language-side phenomenon.}
The Hanson--Kuhn extension of Wittgenstein's distinction, that observation is theory-laden, that the conceptual frame fixes what is observed \citep{hanson1958patterns,kuhn1962structure}, predicts that the linguistic context should be load-bearing for which aspect is reported. Our results give that prediction empirical content. The vision-side feature pools coexist on bistable stimuli (\cref{sec:results-superposition}); what changes when we change the prompt is not what the vision encoder represents but what the language model commits to. ``Is this an X or a Y?'' is a minimal theory-frame, and its effect on the report is large enough to push 100\% abstention to 100\% commitment without any change to the input image. The locus of theory-ladenness, in this VLM, is the language decoder. We caveat: the localization is verified for one VLM and a single SAE training run; cross-model and cross-SAE generalization, as well as random-feature and permuted-aspect baselines, are left to extended work.


\bibliographystyle{icml2026}
\bibliography{references}

\appendix
\section{Dataset inventory}
\label{app:dataset}

The full bistable stimulus set (\cref{sec:methods}) contains 83 images: 29 Panagopoulou benchmark images \citep{panagopoulou2024aspect}, 53 \textsc{AmbiBench} bistable images \citep{ambibench2026}, and 1 programmatically rendered Necker cube. \cref{tab:groups} summarizes the six analysis groups. Pure-aspect controls are SDXL-generated for the photographic groups (\texttt{duck\_rabbit} ducks/rabbits in pond and field settings, \texttt{young\_old\_woman} portraits, etc.) and programmatic for the geometric groups; every control image was hand-verified by the authors and rejected if the non-target aspect was visible or the subject was ambiguous. Public-domain originals from Wikimedia Commons supplement the SDXL-generated controls where available (e.g.\ original Jastrow 1899 duck--rabbit, original Hill 1915 young/old woman). Stimulus IDs are hash-based; full per-image source, license, and retention flags are in the supplementary CSV (\texttt{dataset.csv}).

\begin{table}[h]
  \centering
  \small
  \begin{tabular}{lrrr}
    \toprule
    Group & Bistable & Pure-A & Pure-B \\
    \midrule
    \texttt{duck\_rabbit}      & 12 & 26 & 27 \\
    \texttt{face\_vase}        & 16 & 21 & 27 \\
    \texttt{hidden\_face}      & 15 & 27 & 27 \\
    \texttt{young\_old\_woman} & 8  & 27 & 27 \\
    \texttt{schroeder\_stairs} & 4  & 25 & 8  \\
    \texttt{necker\_cube}      & 14 & 27 & 15 \\
    \bottomrule
  \end{tabular}
  \caption{Per-group stimulus and control counts after curation. Pure-B counts for \texttt{schroeder\_stairs} and \texttt{necker\_cube} are reduced because the canonical pure-B prompts (descending stairs, cube viewed from below) are geometrically ambiguous to SDXL; for \texttt{necker\_cube} we supplement with programmatic 3D solid-cube renders with explicit lighting cues.}
  \label{tab:groups}
\end{table}

\section{Phase 2 feature grids}
\label{app:phase2}

\begin{figure*}[!t]
  \centering
  \includegraphics[width=0.95\textwidth]{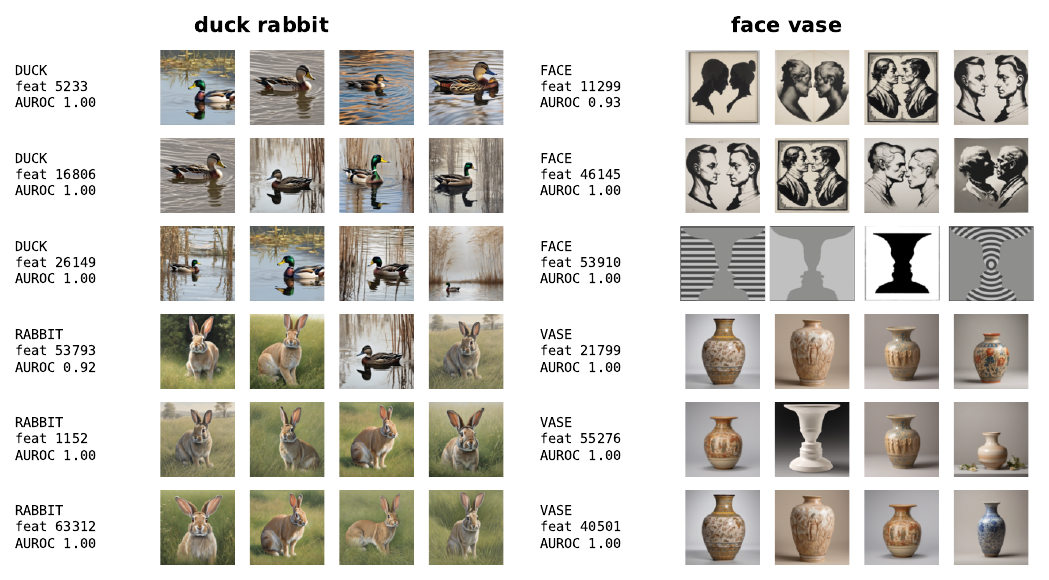}
  \caption{\texttt{duck\_rabbit} and \texttt{face\_vase} retained SAE features. Top-3 features per aspect, with top-4 max-activating images across controls and bistable stimuli. Duck-features fire on duck controls, rabbit-features on rabbit controls; face-features fire on profile controls and the canonical Rubin bistable (face-leaning), vase-features on vase controls.}
  \label{fig:features-app}
\end{figure*}

\cref{fig:features-app} shows the compact two-group view referenced from the main text (\cref{sec:results-features}): top-3 features per aspect for \texttt{duck\_rabbit} and \texttt{face\_vase}, each with top-4 max-activating images. The per-group full grids (15 features per aspect, top-10 images each) for all six groups are in \cref{fig:app-hidden-face,fig:app-young-old,fig:app-schroeder,fig:app-necker} below; equivalent full grids for \texttt{duck\_rabbit} and \texttt{face\_vase} accompany this submission in the supplementary figure bundle. Median leave-one-out tie-corrected AUROC is $\geq 0.997$ for every group; median CC3M-distractor AUROC is $\geq 0.997$ for every group except \texttt{schroeder\_stairs} (0.999) and \texttt{hidden\_face} (0.998). For \texttt{schroeder\_stairs} the pure-B controls are thinner ($n=8$); the AUROC remains high because the controls that survive curation are clean, but the small sample size means per-feature claims are weaker than for the photographic groups.

\begin{figure*}[!p]
  \centering
  \begin{subfigure}{0.48\textwidth}
    \centering
    \includegraphics[width=\textwidth, trim={0 0 0 400pt}, clip]{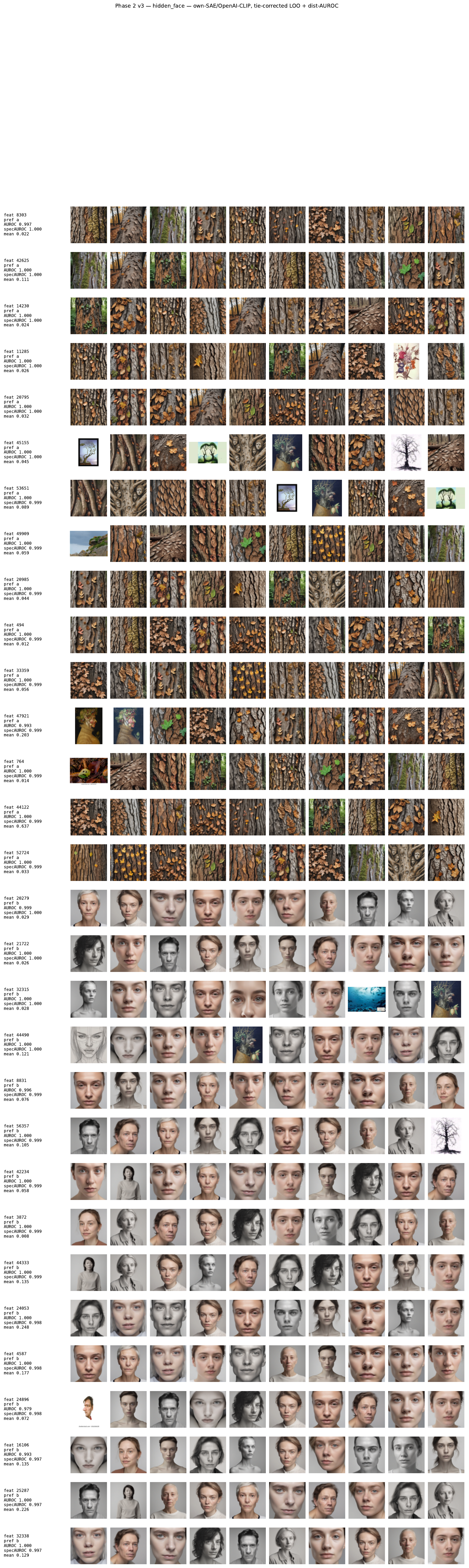}
    \caption{\texttt{hidden\_face}: 15 face-features (top) and 15 background-tree-features (bottom).}
    \label{fig:app-hidden-face}
  \end{subfigure}
  \hfill
  \begin{subfigure}{0.48\textwidth}
    \centering
    \includegraphics[width=\textwidth, trim={0 0 0 400pt}, clip]{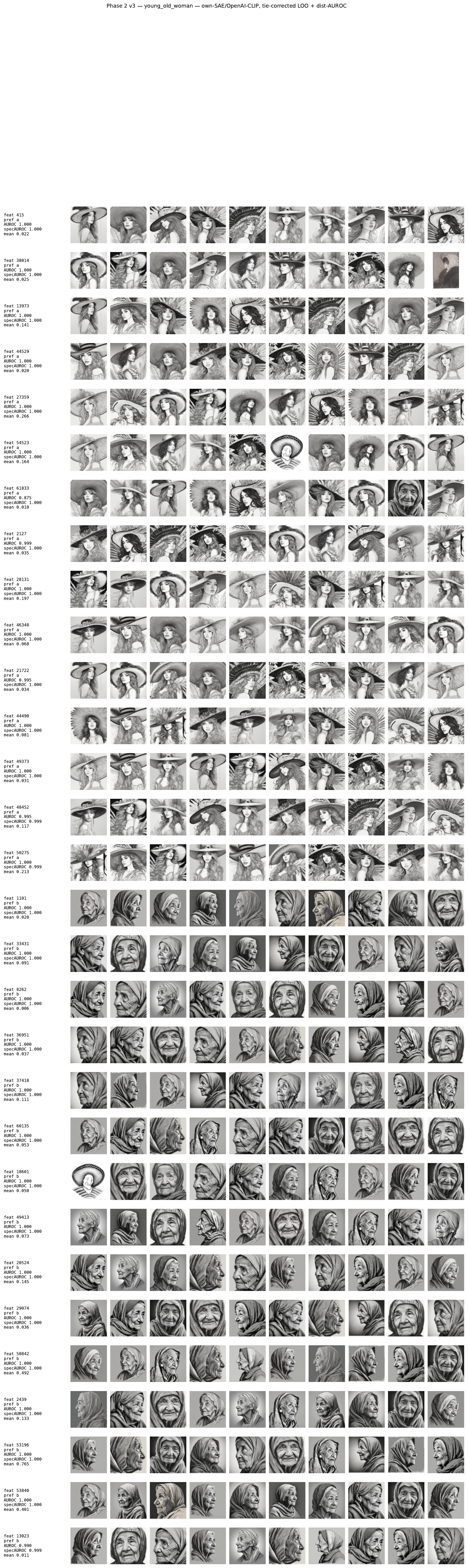}
    \caption{\texttt{young\_old\_woman}: 15 young-features (top) and 15 old-features (bottom).}
    \label{fig:app-young-old}
  \end{subfigure}
\end{figure*}

\begin{figure*}[!p]
  \centering
  \begin{subfigure}{0.48\textwidth}
    \centering
    \includegraphics[width=\textwidth, trim={0 0 0 400pt}, clip]{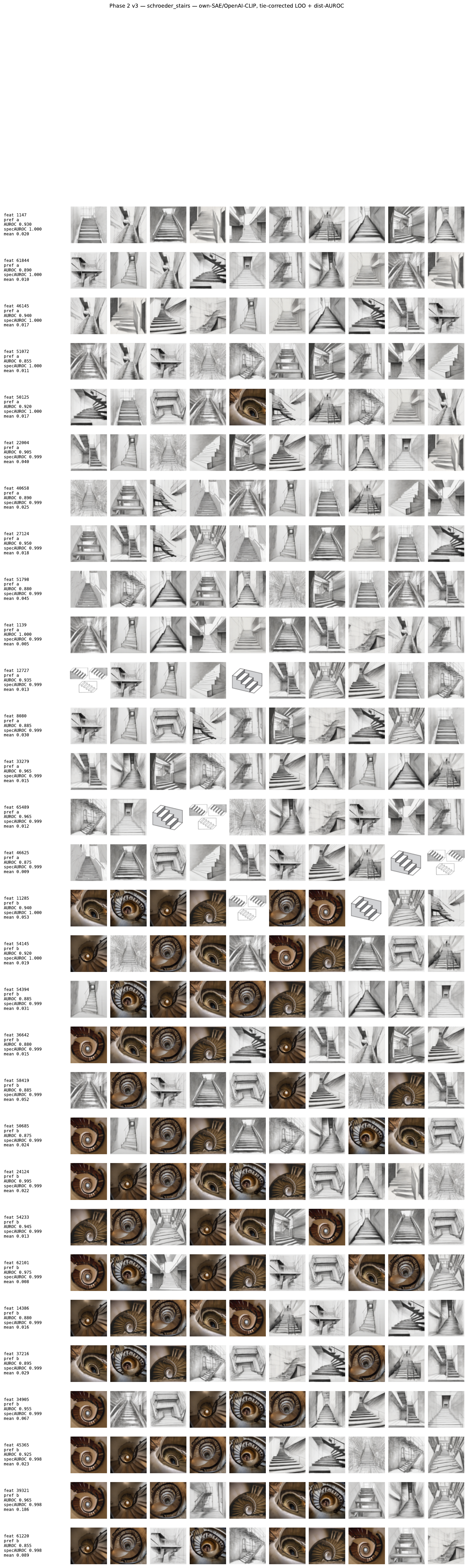}
    \caption{\texttt{schroeder\_stairs}: 15 ascending-features (top) and 15 descending-features (bottom). Pure-B count is $n=8$ post-curation; per-feature evidence is correspondingly thinner.}
    \label{fig:app-schroeder}
  \end{subfigure}
  \hfill
  \begin{subfigure}{0.48\textwidth}
    \centering
    \includegraphics[width=\textwidth, trim={0 0 0 400pt}, clip]{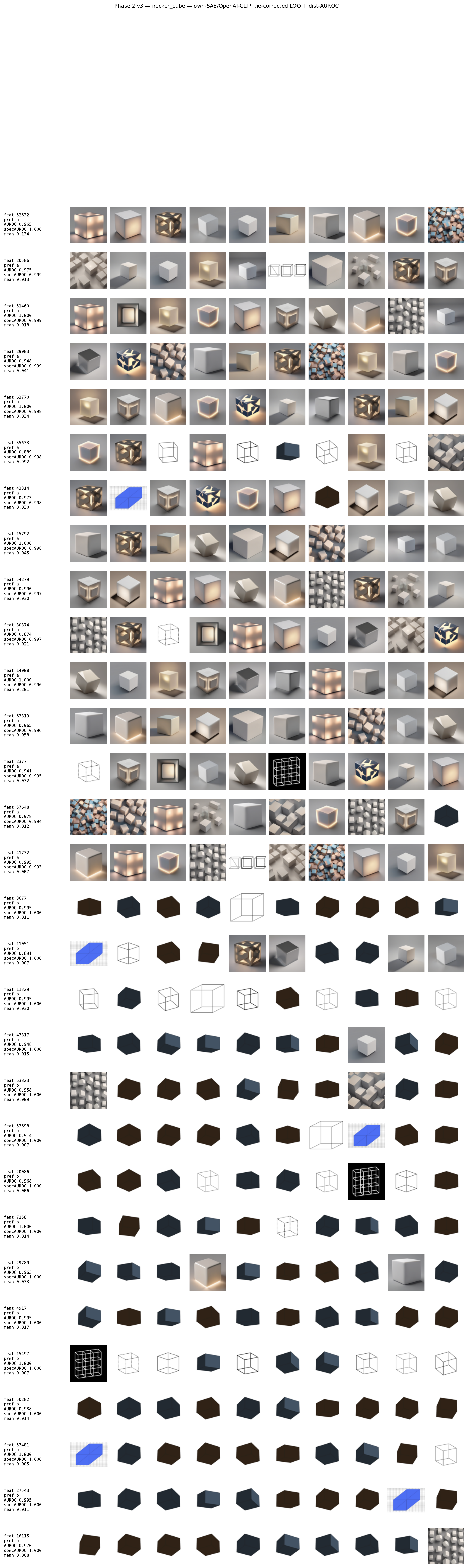}
    \caption{\texttt{necker\_cube}: 15 face-up-features (top) and 15 face-down-features (bottom).}
    \label{fig:app-necker}
  \end{subfigure}
\end{figure*}
\noindent
\begin{minipage}{\columnwidth}
\centering
\includegraphics[width=\columnwidth]{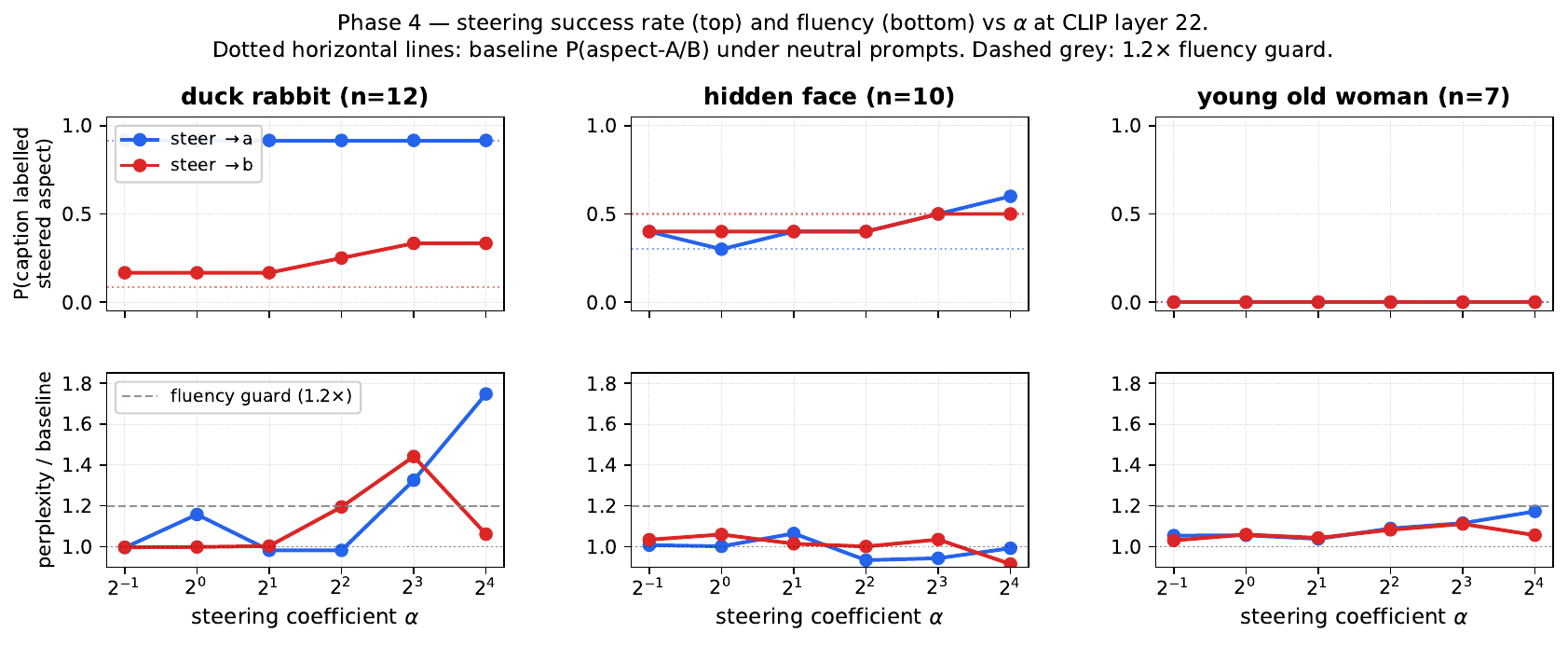}
\captionof{figure}{Phase 4: steering success (top) and fluency (bottom) vs $\alpha$. Vision-side steering rescues default-dominant captions (\texttt{duck\_rabbit}, 33\%) and partially the mixed case (\texttt{hidden\_face}, 50--60\%) under the 1.2$\times$ fluency guard, but \emph{cannot} flip force-balanced captions (\texttt{young\_old\_woman}, 0/7).}
\label{fig:steering}
\end{minipage}
\vspace{1em}

\section{Phase 4 full per-$\alpha$ steering results}
\label{app:phase4}

\cref{tab:phase4} reports the per-direction-per-$\alpha$ flip rate and perplexity ratio for the three steering groups (extracted from \texttt{success\_vs\_alpha.csv}). The fluency guard is a perplexity ratio of 1.2 against the unsteered caption.

\begin{table}[h]
  \centering
  \scriptsize
  \begin{tabular}{lllrrr}
    \toprule
    Group & Dir & $\alpha$ & $n$ & Success & ppl ratio \\
    \midrule
    \texttt{duck\_rabbit}      & a & 0.5  & 12 & 0.92 & 1.00 \\
    \texttt{duck\_rabbit}      & a & 1    & 12 & 0.92 & 1.16 \\
    \texttt{duck\_rabbit}      & a & 2    & 12 & 0.92 & 0.98 \\
    \texttt{duck\_rabbit}      & a & 4    & 12 & 0.92 & 0.98 \\
    \texttt{duck\_rabbit}      & a & 8    & 12 & 0.92 & 1.33 \\
    \texttt{duck\_rabbit}      & a & 16   & 12 & 0.92 & 1.75 \\
    \texttt{duck\_rabbit}      & b & 0.5  & 12 & 0.17 & 1.00 \\
    \texttt{duck\_rabbit}      & b & 1    & 12 & 0.17 & 1.00 \\
    \texttt{duck\_rabbit}      & b & 2    & 12 & 0.17 & 1.00 \\
    \texttt{duck\_rabbit}      & b & 4    & 12 & 0.25 & 1.19 \\
    \texttt{duck\_rabbit}      & b & 8    & 12 & 0.33 & 1.44 \\
    \texttt{duck\_rabbit}      & b & 16   & 12 & 0.33 & 1.06 \\
    \midrule
    \texttt{hidden\_face}      & a & 0.5  & 10 & 0.40 & 1.01 \\
    \texttt{hidden\_face}      & a & 1    & 10 & 0.30 & 1.00 \\
    \texttt{hidden\_face}      & a & 2    & 10 & 0.40 & 1.06 \\
    \texttt{hidden\_face}      & a & 4    & 10 & 0.40 & 0.93 \\
    \texttt{hidden\_face}      & a & 8    & 10 & 0.50 & 0.94 \\
    \texttt{hidden\_face}      & a & 16   & 10 & 0.60 & 0.99 \\
    \texttt{hidden\_face}      & b & 0.5  & 10 & 0.40 & 1.03 \\
    \texttt{hidden\_face}      & b & 1    & 10 & 0.40 & 1.06 \\
    \texttt{hidden\_face}      & b & 2    & 10 & 0.40 & 1.01 \\
    \texttt{hidden\_face}      & b & 4    & 10 & 0.40 & 1.00 \\
    \texttt{hidden\_face}      & b & 8    & 10 & 0.50 & 1.04 \\
    \texttt{hidden\_face}      & b & 16   & 10 & 0.50 & 0.92 \\
    \midrule
    \texttt{young\_old\_woman} & a & 0.5  & 7  & 0.00 & 1.05 \\
    \texttt{young\_old\_woman} & a & 1    & 7  & 0.00 & 1.06 \\
    \texttt{young\_old\_woman} & a & 2    & 7  & 0.00 & 1.04 \\
    \texttt{young\_old\_woman} & a & 4    & 7  & 0.00 & 1.09 \\
    \texttt{young\_old\_woman} & a & 8    & 7  & 0.00 & 1.12 \\
    \texttt{young\_old\_woman} & a & 16   & 7  & 0.00 & 1.17 \\
    \texttt{young\_old\_woman} & b & 0.5  & 7  & 0.00 & 1.03 \\
    \texttt{young\_old\_woman} & b & 1    & 7  & 0.00 & 1.06 \\
    \texttt{young\_old\_woman} & b & 2    & 7  & 0.00 & 1.04 \\
    \texttt{young\_old\_woman} & b & 4    & 7  & 0.00 & 1.08 \\
    \texttt{young\_old\_woman} & b & 8    & 7  & 0.00 & 1.11 \\
    \texttt{young\_old\_woman} & b & 16   & 7  & 0.00 & 1.06 \\
    \bottomrule
  \end{tabular}
  \caption{Phase 4 per-$\alpha$ flip rate and perplexity ratio. ``Dir~a'' steers toward the canonical aspect-A; ``Dir~b'' toward aspect-B. The \texttt{young\_old\_woman} sample is the seven Phase-3 superposition stimuli; \texttt{pana\_028} (Phase-3 \texttt{dominance\_b}) is excluded by the protocol's superposition filter.}
  \label{tab:phase4}
\end{table}

The \texttt{young\_old\_woman} captions remain aspect-agnostic across the entire $\alpha$ range, then degrade fluency as $\alpha$ grows without ever producing an aspect-committed caption: at $\alpha=16$ the captions describe ``a stylized illustration of a woman's face with a dramatic, exaggerated eyelash and a large flowing hair accessory that resembles a feather or a piece of fabric'' --- the steering direction influences low-level visual descriptors (hair, hat) but the model continues to refuse to commit to ``young'' or ``old.''
\section{Rank-tie diagnostic}
\label{app:rank-tie}

This appendix reports the empirical signature that motivated the tie-corrected ranking note in \cref{sec:methods}. An earlier feature-identification pass used a vectorized AUROC built on \texttt{numpy.argsort} and \texttt{np.put\_along\_axis} for ranking. \texttt{numpy.argsort} is a stable sort: when two activations are equal, it preserves their original row order. In our data, class-A controls occupy rows $0\ldots n_A-1$ of the activation matrix and class-B controls occupy rows $n_A\ldots n_A+n_B-1$, so a value of zero shared between an A row and a B row gets a lower rank when it is in an A row, and a higher rank when it is in a B row, purely by virtue of position.

Because TopK SAE outputs are over 99\% sparse, most feature columns contain mass at exactly zero. The argsort tie-break therefore systematically gave class-A samples low ranks and class-B samples high ranks on any column with no real signal, producing AUROCs near zero (interpreted as ``B-preferring'') for thousands of features that in fact had no class-discriminative behavior. The empirical signature was striking: across all six analysis groups, the post-AUROC candidate counts were asymmetric by 47--79$\times$ in favor of phantom B-preferring features (\cref{tab:rank-tie}), and the retained B-features clustered in a narrow feature-index range (43000--44200) consistent across groups, with mean activations $\sim 0$ on matching controls.

\begin{table}[h]
  \centering
  \small
  \begin{tabular}{lrrr}
    \toprule
    Group & Phantom-A & Phantom-B & Ratio \\
    \midrule
    \texttt{duck\_rabbit}      & 208 & 9{,}848  & 47$\times$ \\
    \texttt{face\_vase}        & 180 & 14{,}253 & 79$\times$ \\
    \texttt{hidden\_face}      & 297 & 14{,}636 & 49$\times$ \\
    \texttt{young\_old\_woman} & 272 & 12{,}548 & 46$\times$ \\
    \texttt{schroeder\_stairs} & 121 & 9{,}551  & 79$\times$ \\
    \texttt{necker\_cube}      & 138 & 9{,}578  & 69$\times$ \\
    \bottomrule
  \end{tabular}
  \caption{Phantom feature counts, pre-fix. Each group's class-B candidate set is dominated by features with no real activation but spurious AUROC near 0 induced by stable-sort tie-breaking on row position.}
  \label{tab:rank-tie}
\end{table}

The fix is one line: \texttt{scipy.stats.rankdata(method
="average", axis=0)} replaces the argsort-based ranking and assigns each tied set the average of the rank positions it spans. With the tie-corrected ranking, the asymmetry disappears: post-AUROC candidate counts equalize, the phantom B-features vanish, and the surviving features in both classes have substantive mean activation on matching controls and clear max-activating-image evidence (cf.\ \cref{fig:features-app}). We also added a sparsity floor (\texttt{mean\_match}~$>0.005$) and replaced the previous specificity ratio with a CC3M distractor AUROC; together these changes leave 15 features per aspect per group from a starting pool of 65{,}536.

We flag the failure mode because it is silent. \texttt{numpy.argsort} produces no warning on ties, AUROC values look reasonable on inspection (they are not NaN, they are bounded in $[0,1]$, and many appear close to 1 because the phantom B-preference is systematic), and the only diagnostic that catches it is checking max-activating images by hand. We expect this pitfall to recur as more researchers analyze SAE features with rank statistics, and we recommend tie-corrected ranking as a default in any AUROC pipeline that consumes sparse activations.

\end{document}